# Arabic Call system based on pedagogically indexed text

Mohamed Achraf Ben Mohamed [2,3], Dhaou El Ghoul [1], Mohamed Amine Nahdi [1], Mourad Mars [1,2]
Mounir Zrigui [2,3]
[1] LIDILEM, University of Stendhal, Grenoble3, France
[2] UTIC Laboratory, University of Monastir, Tunisia
[3] Faculty of sciences of Monastir, Tunisia

**Abstract** - *This article introduces the benefits of using computer as a tool for foreign language teaching and learning. It describes the effect of using Natural Language Processing (NLP) tools for learning Arabic. The technique explored in this particular case is the employment of pedagogically indexed corpora. This text-based method provides the teacher the advantage of building activities based on texts adapted to a particular pedagogical situation. This paper also presents ARAC: a Platform dedicated to language educators allowing them to create activities within their own pedagogical area of interest.*

## 1 Introduction

Computer-assisted language learning (CALL) software emerged since the 1960's and have been making continuous progress, especially in the recent years with the vulgarization of new technologies and the development of new teaching methods. There are three distinct phases in the evolution of CALL [14]:

- Communicative CALL (1970's-1980's) was based on the behaviorist theories of learning and teaching, this approach is based on drills and repetition.

- Structural CALL (1980's-1990's) rejected behaviorist approach and insisted that CALL should focus more on using forms rather than on the forms themselves.

- Integrative CALL (21st century) is based on communicative CALL with the integration of two advanced technologies into language teaching: multimedia, on one hand and the development of networks and the Internet on the other.

## 2 Motivations behind CALL emergence

Since computers became widespread, education became more concrete and more assimilated by learners who are no longer just passive listeners, especially in language learning, rather as active participants [5, 10, 13]. In the following we are going to detail the advantages of CALL and we will justify the huge potential it has acquired.

### 2.1 Learner autonomy

Teaching based on new technologies like the Internet, can help learners improve their language skills by helping them develop strategies for self-learning and promote self-confidence. Student motivation is increased, especially when a variety of activities is proposed, which makes them feel more independent [1].

### 2.2 Authentic learning

The computer can improve the authenticity of the language enabling learners to see and hear native speakers interacting in the language, not only verbal but also devoted wholly or partially in the facial expressions and gestures [14]. The Internet also facilitates the use of a domain specific language [12]. It is an endless source of authentic materials. Learners will have the opportunity to use various resources of authentic materials which incorporate graphics, audio and video, students can improve their ability to speak, to increase their comprehension skills, expand their vocabularies, to test their pronunciation, and practice their reading and writing.

### 2.3 Personalized learning

Timid or fearful learners can be greatly beneficiaries of this individualized learning centered on learner. 'Fast' learners can also reach their full potential without preventing their peers to work at their own rhythm. In language learning, teachers are confronted by several skills and learning capacities [7]. Computers can help teachers to 'orchestrate' different rhythms of different learners by giving them adaptable learning methods and tools. This encourages learners to take more responsibility for their own learning.

### 2.4 Benefits of using multimedia

The use of computers in language teaching has come to break with the old methods of learning that were passive and boring. Computers have helped to integrate photos and video. Previously considered abstract concepts become, through simulation for example, more real and understandable and learners can study more actively [4, 9].

## 2.5 Learning beyond the limits of time and space

Through computer networks, learners are now able to access knowledge wherever and whenever they want, not only that, particularly the web has democratized learning by making it accessible to everyone regardless of their ethnicity, origin or income [12]. In addition the web offers the opportunity to create learning groups including learners and teachers.

## 3 Pedagogical text indexation

### 3.1 Definition

«Pedagogical indexation is an indexation performed according to a documentary language that allows the user to search for objects to use in teaching. » [7].

Our center of interest is language learning and our aim is mainly educational, in another way the expectation of the user (teacher) of a document indexed for learning would be to extract objects that can be used as part of its course (Arabic learning in our case) with an application of CALL, this implies that this base of indexed texts must ensure, through its interface, the following features [6, 8]:

- Enabling teachers to make requests based on criteria consistent with the teaching of languages.

- Allow teachers to add their own texts to the base.

- Language teachers are not necessarily very knowledgeable in computers; the interface should be as friendly as possible.

### 3.2 Educational Metadata Standards

Educational Metadata Standards are initiatives in the field of standards for e-Learning and information and communication technology (ICT) designed to describe objects by metadata [9]. The term metadata is used to define all the technical and descriptive information added to documents to better describe them. Metadata were originally proposed to improve Web search, they are to enrich the documents published by a set of information to better identify them: name of the author and publisher, publication date, Summary of content, etc...

The oldest standard concerned with the description of learning objects is the Dublin Core Metadata Initiative (DCMI), there are other initiatives resulting directly from DCMI, Getaway to Educational Material (GEM) and Educational Network of Australia (EdNA), however, the most common standard used today is Learning Object Metadata (LOM) [9]. For the description of texts constituting the corpora of our platform ARAC, we pick up the LOM standard.

## 4 ARAC Platform

### 4.1 Presentation

Our modules-based platform ARAC (ARAbic Call) has been developed to allow users to manage and exploit a corpus from a single interface with tools both powerful and easy to use to automate the work needed. The platform ARAC was developed in an ASP mode (Application Service Provider); this mode avoids the technical constraints and allows users to automatically benefit from any updates. Generally, only a password and login are required to access the platform at any time (Authentication). This technology (ASP) requires no special configuration and therefore excludes any costs of installation, operation or maintenance engendered by the most heavy software solutions. Throughout the development of this platform we have tried to make the technical side transparent from the user (teacher) who should only focus on the educational aspect.

### 4.2 Architecture

ARAC platform consists of four modules with high synergy; it was developed in accordance with the general functional diagram of learning system [2, 3, 11] (Figure 1).

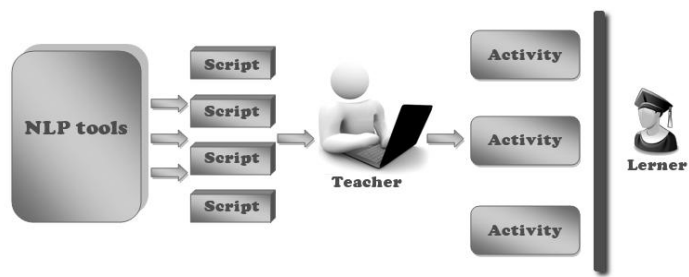

Figure 1 - General functional diagram of learning system

Figure 2 presents the architecture of our platform; in the following we will give an overview of each module.

#### 4.2.1 Text management module

This is the input module of the platform by texts encoded using the standard UTF-8. The choice of this coding was justified by the fact that, firstly, most of the Arab digital textual resources are encoded using this standard, and secondly because the standard UTF-8 is supported by the most popular browsers. [18] The management module corpus is responsible from two tasks, namely: classification and annotation.

- Classification: The corpus is composed of several themes created by the administrator and each containing texts. Very often the teacher is looking for a text in a particular subject.

- Annotation: ARAC platform offers two forms of annotation, the first is automatic (figure 3), it allows the user to identify particles such as particles of coordination or interrogative particles, the second is manual, it allows the user to identify any other lexemes [17] [19].

The algorithm of automatic annotation (Annotatik) is shown in the table 1.

Table 1- The Pseudo code for automatic annotation algorithm.

| **Algorithm** Annotatik (F, T) |
|---|
| **Input**: File F , T // F:Corpora, T:taxonomies |
| **Output**: Nothing (but insert in the database) |
| Lock file(F) // *Prevent concurrent file access.* <br> lexemes[] ← split file content (F) //*retrieve the lexemes in a table.* <br> taxonomies[]← split file content (T) //*retrieve the taxonomies in a table.* <br> counter ← 0 <br> rasult[] = array <br> **for**(j=0; j<count(taxonomies); ++j){ <br>   **for**(i=0; i<count(lexemes); ++i){ <br>     **if**(taxonomies[j] == lexemes[i] ){ // *looks for elements of the taxonomies table present in the text.* <br>       result [counter ++] = i; // *group results.* <br>     } <br>   } <br> } <br><br> **if**( count(result) > 0) { // *at least one element has been found.* <br>   **for**(k=0; k<count(result); ++k) <br>     insert into the database (result[k]) |

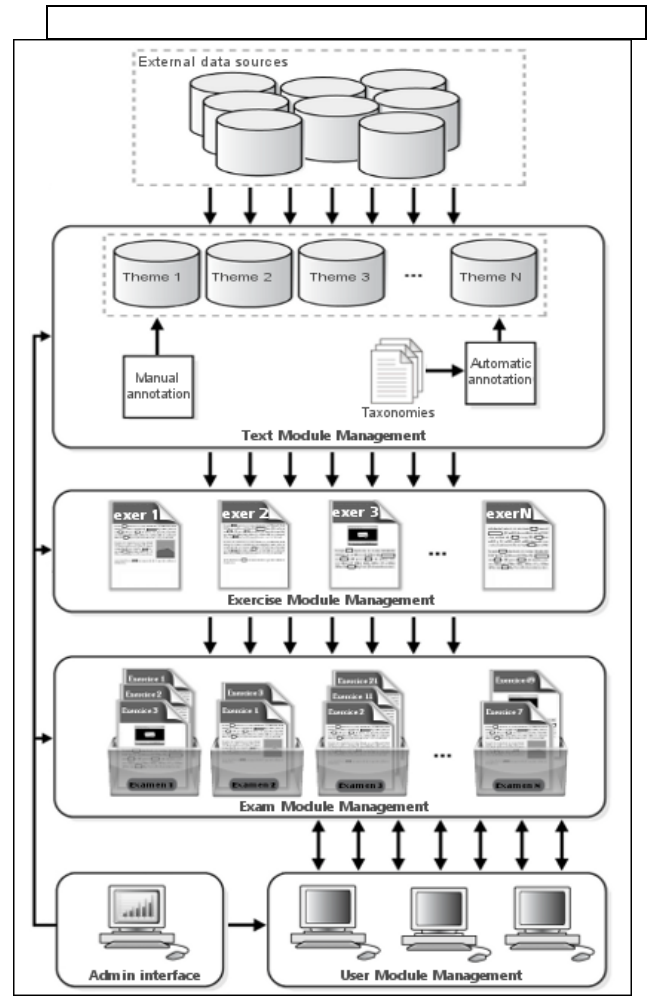

Figure 2 - Architecture of the Platform ARAC

#### 4.2.2 Exercise management

With this module, we move from the acquisition of text step to the exploitation of text step. The production of an exercise requires two steps: first, the choice of text responding to the request of the teacher, and second, formatting the exercise using a WYSIWYG editor (figure 4).

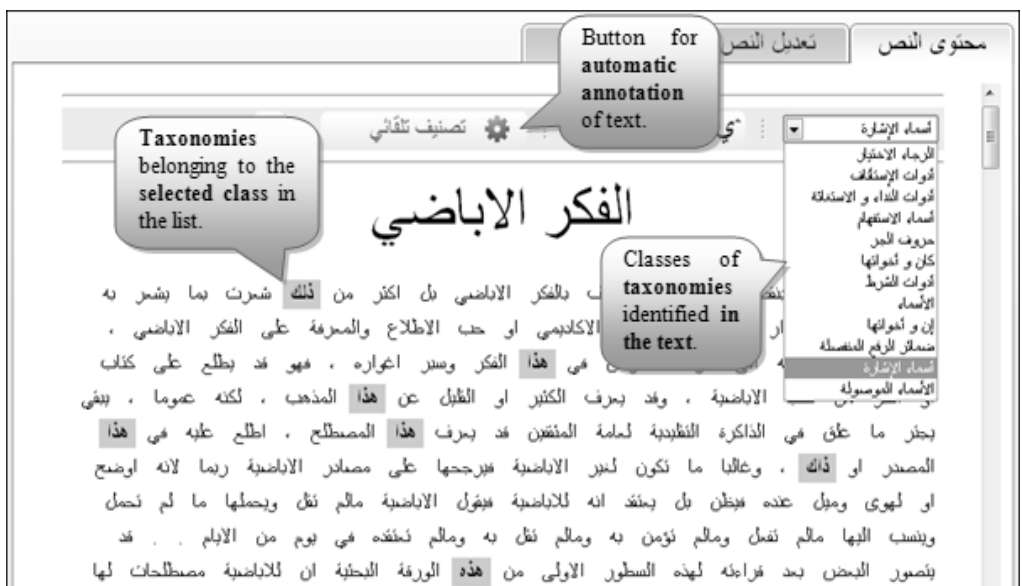
Figure 3 - Automatic annotation

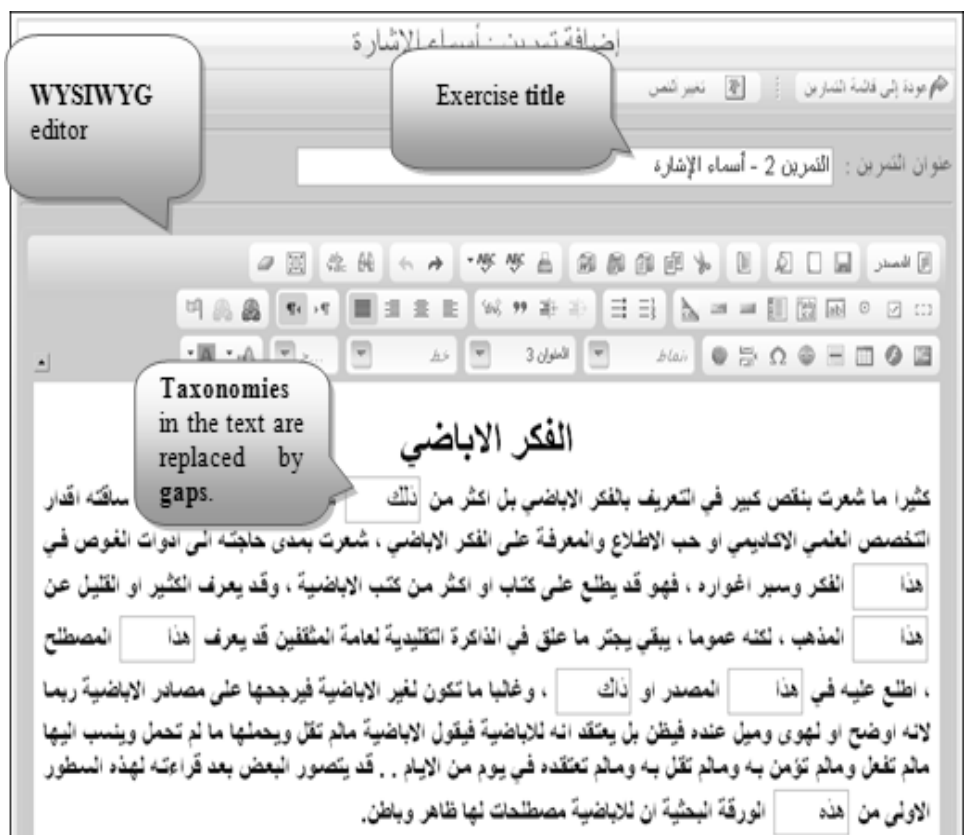
Figure 4 - Exercise formatting.

### 4.2.3 Exam management

An exam is an ordered selection of exercises selected from the available exercises. After being created, the exam will be assigned to students by the teacher. Figure 5 shows the final overview of the activity.

Figure 5 - Preview of a non accomplished exam

### 4.2.4 User Management

User management covers several aspects such as user account creation, exams monitoring (accomplished or not), the performances (Figure 7) and deleting accounts.

Performance is given by the following formula:

$$Performance = \frac{Number\ of\ Correct\ Answers}{Number\ of\ Questions} \times 100 \qquad (1)$$

### 4.3 Example of activity

For each activity, the learner will be asked to complete the empty boxes with appropriate terms, and once the form is validated, the student will receive a page containing the answers to the exam and the results are verified (Figure 6). This correction page will indicate, by a set of colors, where the learner has done good (green) or bad (red) responses. In addition to that, the correction will be provided with statistics.

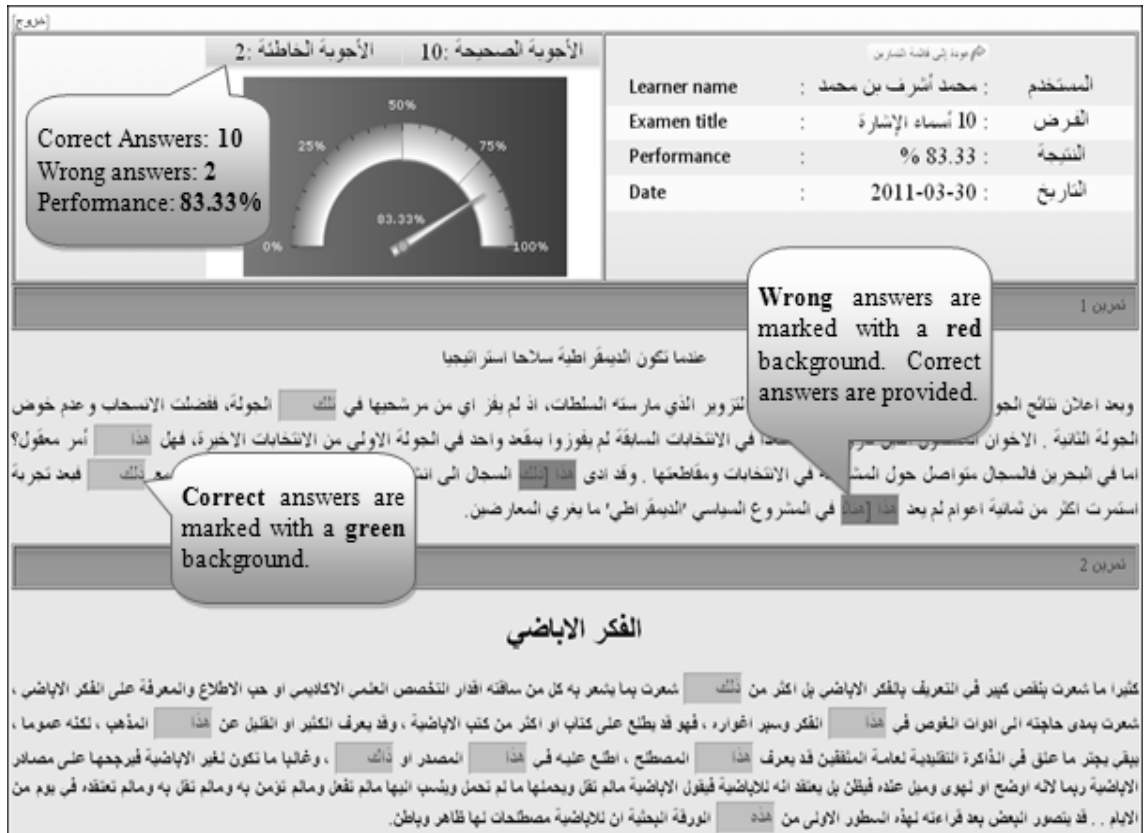

Figure 6 - Preview of an accomplished exam.

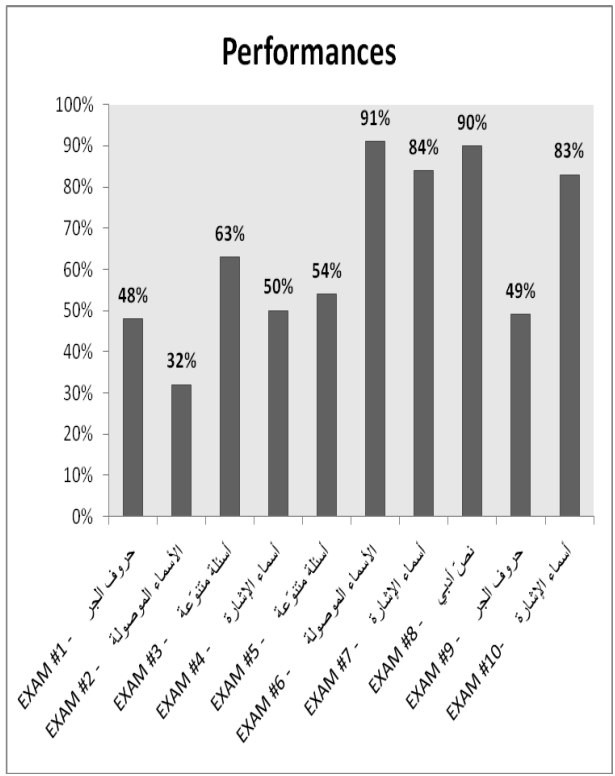

Figure 7 - Performances achieved by a student

## 5 Conclusions

We developed a tool [16] for creating and exploiting a database of text indexed pedagogically; this tool will allow language teachers to create activities within their own pedagogical area of interest. We chose to develop the ARAC platform in ASP mode, thus making the implementation and updating of this tool as simple as possible and also to ensure full mobility of data access whether to the teacher or learner, independently of the machine and operating system in use (platform independent).

For future work we plan to improve our platform ARAC in some aspects, such as the possibility of treating any type of data (e.g. RSS), we can also consider automating the classification, we also plan to add an analysis and feedback module.